\documentclass[sigconf]{acmart}

\pdfoutput=1
\renewcommand\footnotetextcopyrightpermission[1]{} 
\usepackage{booktabs}
\usepackage[ruled,linesnumbered,vlined]{algorithm2e}
\usepackage{epstopdf}
\usepackage{graphicx,subfigure}

\begin{document}

\title{Competitive Bridge Bidding with Deep Neural Networks}  
\titlenote{This paper was submitted to AAMAS on Nov. 12, 2018, accepted on Jan. 23, 2019.}


\author{Jiang Rong}
\affiliation{%
  \institution{Institute of Computing Technology, Chinese Academy of Sciences}
  \institution{University of Chinese Academy of Sciences, Beijing, 100190}
}
\email{jiangrong.hb@gmail.com}

\author{Tao Qin}
\affiliation{%
  \institution{Microsoft Research}
  \city{Beijing, 100080} 
}
\email{taoqin@microsoft.com}

\author{Bo An}
\affiliation{%
	\institution{School of Computer Science and Engineering, Nanyang Technological University}
	\city{Singapore, 639798} 
}
\email{boan@@ntu.edu.sg}

\begin{abstract}  
The game of bridge consists of two stages: bidding and playing. While playing is proved to be relatively easy for computer programs, bidding is very challenging. During the bidding stage, each player knowing only his/her own cards needs to exchange information with his/her partner and interfere with opponents at the same time. Existing methods for solving perfect-information games cannot be directly applied to bidding. Most bridge programs are based on human-designed rules, which, however, cannot cover all situations and are usually ambiguous and even conflicting with each other. In this paper, we, for the first time, propose a competitive bidding system based on deep learning techniques, which exhibits two novelties. First, we design a compact representation to encode the private and public information available to a player for bidding. Second, based on the analysis of the impact of other players' unknown cards on one's final rewards, we design two neural networks to deal with imperfect information, the first one inferring the cards of the partner and the second one taking the outputs of the first one as part of its input to select a bid. Experimental results show that our bidding system outperforms the top rule-based program.
\end{abstract}

%


\maketitle


\section{Introduction}
Games have long been of great interest for artificial intelligence (AI) researchers. One set of works focus on full-information competitive games such as chess \cite{grammenos2005ua} and go \cite{coulom2007computing,silver2016mastering}. Such games present two challenges: the large state space for decision-making and the competition from the opponent player. The AI program AlphaGo \cite{silver2016mastering,silver2017mastering} has achieved great success in the game of Go. The other set of AI researches investigate imperfect-information card games, such as poker \cite{sandholm2010state,yakovenko2016poker} and bridge \cite{amit2006learning,ginsberg2001gib,yeh2016automatic}. The computer programs Libratus \cite{brown2017libratus} and DeepStack \cite{moravvcik2017deepstack} for no-limit Texas hold'em both showed expert-level performance, but their techniques can only handle the heads-up (two-player) situation. 

Contract bridge, or simply bridge, is one of the most interesting and difficult card games, because 1) it presents all the above challenges, i.e., large state space, competition and imperfect information; 2) four players rather than two in bridge makes the methods designed for two-player zero-sum games (e.g., heads-up no-limit poker \cite{sandholm2010state,yakovenko2016poker}), which focus on Nash equilibrium finding/approximation, not applicable; 3) it has to deal with cooperation between partners. 
The best bridge AI-programs, such as GIB\footnote{GIB. \url{http://www.gibware.com/}}, Jack\footnote{Jack. \url{http://www.jackbridge.com/}} and Wbridge5\footnote{Wbridge5. \url{http://www.wbridge5.com/}}, have not yet reached the level of top human experts, which is probably because of the weakness of their bidding systems \cite{amit2006learning,ginsberg2001gib}.

The game of bridge consists of two parts, bidding and playing.  Playing is relatively easy for AI agents and many programs (e.g., GIB, Jack and Wbridge5) have shown good playing performance \cite{ amit2006learning,ginsberg2001gib}. For example, in 1998, the GIB program attained the 12th place among 35 human experts in a contest  without bidding, which demonstrates that computer bridge agents can compete against human expert players in the playing stage. In human world championships, the variation in the level of the players during card playing is also negligible, making the quality of the bidding the decisive factor in the game \cite{amit2006learning}.

Bidding is the hardest part of bridge. During bidding, the players can only see their own cards and the historical bids and try to search for a best contract together with their partners. The difficulty arises from the imperfect-information setting and the complex meanings of the bids. A bid can carry one or more of the following purposes: 1) suggesting an optional contract, 2) exchanging information between partners and 3) interfering with opponents. Human players design a large number of complex bidding rules to explain the meanings of bidding sequences and then suggest a bid based on one's own cards. To the best of our knowledge, almost all bridge programs are based on such human-designed rules. However, since there are $6.35\times 10^{11}$ possible hand holdings with 13 out of 52 cards \cite{amit2006learning} and $10^{47}$ possible bidding sequences \footnote{See the analysis at \url{http://tedmuller.us/Bridge/Esoterica/CountingBridgeAuctions.htm}}, it is unfeasible for the rules to cover all the situations. Hence, the bidding rules are usually ambiguous. 
Besides, as the bidding rules are hand-crafted, some of them may be inefficient and it is very likely that a pair of hand and bidding sequence is not covered by any rule or satisfies multiple rules suggesting conflicting bids. 

Considering these drawbacks, many researchers study how to improve the rule-based computer bidding systems. Amit and Markovitch \cite{amit2006learning} used Monte Carlo Sampling to resolve the conflicts, but did not consider the ambiguity problem. Some researchers tried to infer the cards of other players on the basis of their calls \cite{amit2006learning,ando2003cooperation,ando2000reasoning,ginsberg1999gib}. However, because the possible cards in others' hands amount to $8.45\times10^{16}$ and the rules are exactitude, the inference may be very inaccurate due to the computing resource and time limit. DeLooze and Downey \cite{delooze2007bridge} introduced the Self-Organizing Map neural network trained with examples from a human bidding system in order to reduce ambiguities, which is shown only to be effective for no trump hands. Recently, deep neural networks have  achieved unprecedented performance in many games, e.g., playing the go \cite{silver2016mastering} and Atari games \cite{mnih2015human}, and have also 
been applied to bridge bidding. Yeh and Lin \cite{yeh2016automatic} used a reinforcement learning algorithm to train a value network with raw data. However, the training for the system is based on a small dataset with randomly generated games and the competition from opponents is not considered (i.e., opponent players were assumed to always ``pass'').

In this paper, we, for the first time, develop a competitive bidding system based on deep neural networks, which combines supervised learning (SL) \cite{mohri2012foundations} from human expert data and reinforcement learning (RL) \cite{mnih2015human,sutton1998reinforcement} from self-play. Our techniques have the following two novelties. First, we design an efficient feature representation for learning, in which the bidding sequence is encoded to a compact 0-1 vector of 318 bits. Second, to deal with partnership bidding and imperfect information (i.e., unknown cards in the other three players' hands), we propose a card estimation neural network (ENN) to infer the  partner's cards and demonstrate by experiments that one's reward highly depends on his/her partner's cards and the opponents' cards are much less important and even not necessary for finding the optimal contract. The ENN outputs a probability distribution over possible cards, which serves as a part of the features of the policy neural network (PNN) to produce a probability distribution over possible bids. Both neural networks are first trained in the SL stage using expert data. Then we design an RL algorithm based on REINFORCE \cite{williams1992simple} to let the system gradually improve its ability and learn its own bidding rules from self-play. The learning procedure needs the final rewards of the game, which are related to the result of bidding and the outcome of playing. We leverage the double dummy analysis (DDA) \footnote{DDA.  \url{http://bridgecomposer.com/DDA.htm}} to directly compute the playing outcome. We show by experiments that DDA is a very good approximation to real expert playing.

We compare our bidding system with the program Wbridge5 \cite{ventos2017boosting}, the champions of World Computer-Bridge Championship 2016 - 2018. Results indicate that our bidding system outperforms Wbridge5 by 0.25 IMP, which is significant because Ventos et al. \cite{ventos2017boosting} show that an improvement of 0.1 IMP can greatly enhance the bidding strength of Wbridge5.

The rest of the paper is organized as follows. We introduce some basic knowledge of bridge game in Section \ref{sec:background}. Our neural network-based bidding system is introduced in Section \ref{sec:NN-based_BS}. The learning algorithms are proposed in Section \ref{sec:lerning_algorithms}. In Section \ref{sec:experiments}, we conduct extensive experiments to evaluate our bidding system. The conclusion is given in the last section.

\section{Background}
\label{sec:background}
In this section, we introduce the bidding, playing and scoring mechanism of the game of bridge.
\subsection{Bridge Bidding}
The game of bridge is played by four players, commonly referred to as North, South, East and West. The players are divided into two opposing partnerships, with North-South against West-East. The game uses a standard deck of 52 cards with 4 suits (club $\clubsuit$, diamond $\diamondsuit$, heart $\heartsuit$ and spade $\spadesuit$), each containing 13 cards from A down to 2. The club and the diamond are called the minor suits, while the other two are major suits. Each player is given 13 cards and one player is designated as the dealer that proposes the first bid (called opening bid). Then the auction proceeds around the table in a clockwise manner. Each player chooses one bid from the following 38 candidates in his/her turn: 
\begin{itemize}
	\item a bid higher than that of his/her right-hand player according to the ordered \emph{contract set}
	\begin{equation}
	\{1\clubsuit,1\diamondsuit,1\heartsuit,1\spadesuit,1NT,2\clubsuit,2\diamondsuit,\ldots,7NT\},
	\end{equation}
	where NT means no trump;
	\item pass;
	\item double a contract bid by the opponents;
	\item redouble if one's or one's partner's bid is doubled
\end{itemize}

We call the bids in the contract set the \emph{contract bids} and the other three bids the \emph{non-contract bids}. During the bidding, a player can only observe his/her own 13 cards and the historical bids. The bidding stage ends when a bid is followed by three consecutive ``passes''.

A contract is a tuple of the level (1-7) and the trump suit ($\clubsuit$, $\diamondsuit$, $\heartsuit$, $\spadesuit$, or NT), and the partnership that bids the highest contract wins the auction. The winners are called the contractors and the other two players are then the defenders. The player from the contractor partnership who first called the trump suit becomes the declarer and his/her partner the dummy. Bridge is a zero-sum game and the contractors need to win at least the level plus 6 tricks in the playing stage to get a positive score (usually referred to as ``make the contract''). For example, a contract of $4\clubsuit$ proposes to win at least $4 + 6 = 10$ tricks in the playing. The tricks higher than the level plus 6 are called the overtakes. In this example, 2 overtakes for the contractors means that they win 12 tricks in total.

There is a large number of bidding systems, for example, the Standard American Yellow Card (SAYC) \footnote{SAYC. \url{https://en.wikipedia.org/wiki/Standard_American}}. The bidding rules of many systems are usually ambiguous and even conflicting. For example, a rule may say that ``do not bid with a balanced hand after an enemy 1NT opening unless you are strong enough to double'', where the definition of ``strong enough'' is ambiguous. The suggested bid of the rule that ``with 4-4 or longer in the minor suits, open 1$\diamondsuit$ and rebid 2$\clubsuit$'' is conflicting with the rule that ``always open your longer minor and never rebid a five-card minor'' for the hand with 5 clubs and 5 diamonds. Human players usually devote much time to practice together to reduce the ambiguities and conflicts.

\subsection{Playing and Scoring}
The playing begins right after the bidding stage, which runs for 13 rounds (tricks). The player sitting at the left side of the declarer plays his/her first card (called opening lead) and then the dummy exposes his/her cards to all players. The playing continues in a clockwise manner. Each of the players puts one card on the table in each round. A player must follow the lead suit if possible or play another suit. The winning hand of the four cards are based on the following rule: if a trump suit is played, the highest card in that suit wins the trick, otherwise, the highest lead suit card wins the trick. During the play, the declarer plays both his cards and the dummy's. The player who wins the trick has the right to lead for the next round.

The scoring depends on the number of tricks taken by the contractors, the final contract and whether the contract is doubled or redoubled. Besides, in bridge, partnerships can be vulnerable which is predetermined before the game begins, increasing the reward for successfully making the contract, but also increasing the penalty for failure. If the contractors win the number of tricks they committed to, they get a positive score and the defenders a negative score, otherwise the positive score is given to the defenders. The most widely used scoring mechanism is the \emph{Duplicate Bridge Scoring} (DBS) \footnote{DBS. \url{http://www.acbl.org//learn_page/how-to-play-bridge/how-to-keep-score/}}, which encourages players to bid higher contract for more bonuses, in addition to the trick points. For example, if a ``game'' (contracts with at least 100 trick points) is made, contractors are awarded a bonus of 300 points if not vulnerable, and 500 points if vulnerable. A larger bonus is won if the contractors make a ``small slam'' or ``grand slam'', a contract of level 6 and level 7 respectively. However, they might face a negative score if they fail, even if they took most of the tricks.

In real-world clubs and tournaments, team competition is popular, where a team usually has four players. Two of the players, playing as a partnership, sit at North-South of one table. The other two players of the same team sit at East-West of a different table. The two partnerships from the opposing team fill the empty spots at the two tables. During the course of the match, exactly the same deal is played at both tables. Then the sum of the duplicate scores from the two partnerships of a team is converted to the International Match Points (IMPs) \footnote{IMP. \url{http://www.acbl.org/learn_page/how-to-play-bridge/how-to-keep-score/teams/}}. The team with higher IMPs wins the match.

\section{Neural Network-Based Bidding System}
\label{sec:NN-based_BS}
Bridge is a teamwork-based game and two players of a partnership adopt the same bidding system for information exchange. The system for human players consists of the predefined rules, which is a set of agreements and understandings assigned to bids and sequences of bids used by a partnership. Each bidding system ascribes a meaning to every possible bid by each player of a partnership, and presents a codified language which allows the players to exchange information about their card holdings. We implement the bidding system by two neural networks, the ENN (estimation neural network) and the PNN (policy neural network), where the ENN is used to estimate the cards in our partner's hands and the PNN is designed for taking actions based on the information we have got. We will show in the Section \ref{sec:DDA} that it is not necessary to estimate opponents' cards because the distribution of the remaining 26 cards between opponents' hands has little effect on the final results.  

In the following two subsections, we first give the definitions of the two networks and then introduce their feature representations.

\subsection{Definitions of ENN and PNN}
The set of players is represented as
\begin{equation}
\mathcal{P} = \{N, E, S, W\},
\end{equation}
where $N$ and $S$ are in a team, so are $E$ and $W$. We use $p^+$ to denote the partner of player $p\in\mathcal{P}$. Let 
\begin{equation}
C=\{\left\langle x_i\right\rangle _{i=1}^{52}|x_i\in\{0,1\}\wedge\sum\limits_{i=1}^{52}x_i = 13\}
\end{equation}
be the set of possible initial cards of a player. Given the cards $c_p\in C$ of player $p\in \mathcal{P}$, $C_{-c_p}$ represents possible initial hands excluding cards in $c_p$. We use $H$ to indicate the set of all bidding histories and let $V$ be the set of vulnerabilities. A player $p$ can infer the cards of $p^+$ based on the information he/she has got, including his/her own cards $c_p\in C$, the public vulnerability $v\in V$ and the bidding history $h\in H$. Specifically, we define
\begin{equation}
D:C\times V \times H \mapsto [0,1]^{52}
\end{equation}
such that the $i$-th component of $D(c_p, v, h)$ is the probability that, in $p$'s belief, $p^+$ were holding card $x_i$. Our PNN is then defined as a neural network
\begin{equation}
\sigma_\theta:C\times V\times H\times [0,1]^{52}\mapsto [0,1]^{38},
\end{equation}
where $\theta$ represents the network parameters. That is, the PNN's features consist of the cards in one's hands, the vulnerability, the bidding history and the estimation of one's partner's cards. Let $\pi(x|y)$ represent the conditional probability of $x$ given $y$ and it follows that
\begin{equation}
\label{eq:D}
D(c_p, v, h) = \sum\limits_{c_{p^+}\in C_{-c_p}}\pi(c_{p^+}|v, B(p^+, h))\cdot c_{p^+},
\end{equation}
where $B(p^+, h)$ is the set of bids called by $p^+$ in the history $h$. In the above equation, $c_p+$ is a vector and $\pi(c_p+|v,B(p^+,h))$, the post probability of $c_p+$, is a scalar. The product of a scalar and a vector means multiplying each component of the vector by the scalar. Given the PNN, theoretically, we can compute $D(c_p, v, h)$ based on Eq.\!~\eqref{eq:D} and Bayes' rule \cite{feller1968introduction}:
\begin{eqnarray}
\pi(c_{p^+}|v, B(p^+, h))\!\!\!\!\!\!&=&\!\!\!\!\!\!\frac{\pi(c_{p^+},v, B(p^+, h))}{\pi(v, B(p^+, h))}\nonumber\\
\!\!\!\!\!\!&=&\!\!\!\!\!\!\frac{\pi(B(p^+, h)|c_{p^+}, v)\pi(c_{p^+}, v)}{\sum_{c^\prime_{p^+}\in C_{-c_p}}\pi(v, B(p^+, h)|c_{p^+}^\prime)\pi(c_{p^+}^\prime)}\nonumber\\
\!\!\!\!\!\!&=&\!\!\!\!\!\!\frac{\pi(B(p^+, h)|c_{p^+}, v)\pi(c_{p^+}, v)}{\sum_{c^\prime_{p^+}\in C_{-c_p}}\pi(B(p^+, h)|c^\prime_{p^+},v)\pi(v|c^\prime_{p^+})\pi(c^\prime_{p^+})} \nonumber\\
\label{eq:Bayes}
\!\!\!\!\!\!&=&\!\!\!\!\!\! \frac{\pi(B(p^+, h)|c_{p^+}, v)}{\sum_{c^\prime_{p^+}\in C_{-c_p}}\pi(B(p^+, h)|c^\prime_{p^+},v)}.
\end{eqnarray}
Further, we have that
\begin{eqnarray}
\label{eq:pi}
&&\!\!\!\!\!\!\!\!\pi(B(p^+, h)|c_{p^+}, v) \nonumber\\
= &&\!\!\!\!\!\!\!\!\prod\limits_{i}\sigma_{\theta}(B_i(p^+, h)|c_{p^+}, v, h_i(p^+), D(c_{p^+}, v, h_i(p^+))),
\end{eqnarray}   
where $B_i(p^+, h)$ is the $i$-th bid of $p^+$ and $h_i(p^+)$ represents the bidding sequence $p^+$ observed when he/she takes his/her $i$-th actions. Substituting Eqs.\!~\eqref{eq:Bayes} and \eqref{eq:pi} into Eq.\!~\eqref{eq:D} leads to the fact that to compute $D(c_p, v, h)$, we need to recursively apply the Bayes' rule. Since the space size of $C_{-c_p}$ is
\begin{equation}
\binom{39}{13} = 8\times 10^9,
\end{equation}
when the length of $h$ is $n$, the time complexity for computing $D(c_p, v, h)$ is 
\begin{equation}
O((8\times 10^9)^{\frac{n}{2}}) = O(8^{0.5n}10^{4.5n}).
\end{equation}
That is, it is impractical to directly use the Bayes' rule to compute $D(\cdot)$. Thus, we propose the neural network ENN
\begin{equation}
\phi_\omega:C\times V \times H \mapsto [0,1]^{52}
\end{equation}
to approximate $D(\cdot)$, where $\omega$ denote the parameters to be learned. 

In our model, the ENN outputs 52 probabilities about the partner’s cards, which are directly fed into the PNN, because in most situations of bridge, it is unnecessary to know the exact 13 cards of one’s partner and the probability distribution is sufficient for the PNN for making decision. For example, to make a specific contract, one may just want to confirm that his/her partner holds at least 4 minor-suit cards, i.e., the sum of the probabilities of minor suits is not less than 4, no matter which minor-suit card is held. 

In the next subsection, we study how the features of one's cards $c_p$, the vulnerability of both partnerships $v$ and the bidding sequence $h$ are represented.

\subsection{Compact Feature Representation}
We use a 52-dimensional 0-1 vector for a player's cards, where a ``1'' in position $i$ means the player has the $i$-th card in the ordered set
\begin{equation}
\{\clubsuit 2-A, \diamondsuit 2-A, \heartsuit 2-A, \spadesuit 2-A\}.
\end{equation}
A 2-dimensional vector is used to indicate the vulnerability with ``00'' for none of vulnerability, ``11'' for both of vulnerability, ``01'' for favorable vulnerability (only the opponent partnership is vulnerable) and ``10'' for unfavorable vulnerability (in contrast to favorable vulnerability). 

There are 38 possible bids, including 35 contract bids, ``pass'', ``double'' and ``redouble''. According to the bidding instructions of bridge, there are at most 8 non-contract bids after each contract bid, i.e., ``pass-pass-double-pass-pass-redouble-pass-pass''. Note that the dealer can begin the bidding with ``pass'', and if the other three players also choose ``pass'' then the game ends immediately without a playing stage. Thus, the maximum bidding length a player need to consider is $3 + (1 + 8)\cdot35 = 318$. Previous work like \cite{yeh2016automatic} represents an individual bid with a one-hot vector of 38 dimensions, which requires a vector with more than ten thousands of dimensions to represent a bidding history. Such a representation is inefficient in computation and it is assumed in \cite{yeh2016automatic} that the maximal length of a bidding sequence is $L$ $(\leq 5)$ to address this problem. However, we observe from the expert data that more than $83\%$ of the sequences have more than 5 bids. We propose a more compact 318-dimensional vector to represent the bidding sequence. 

\begin{figure*}[h]
	\centering
	\includegraphics[width=120mm]{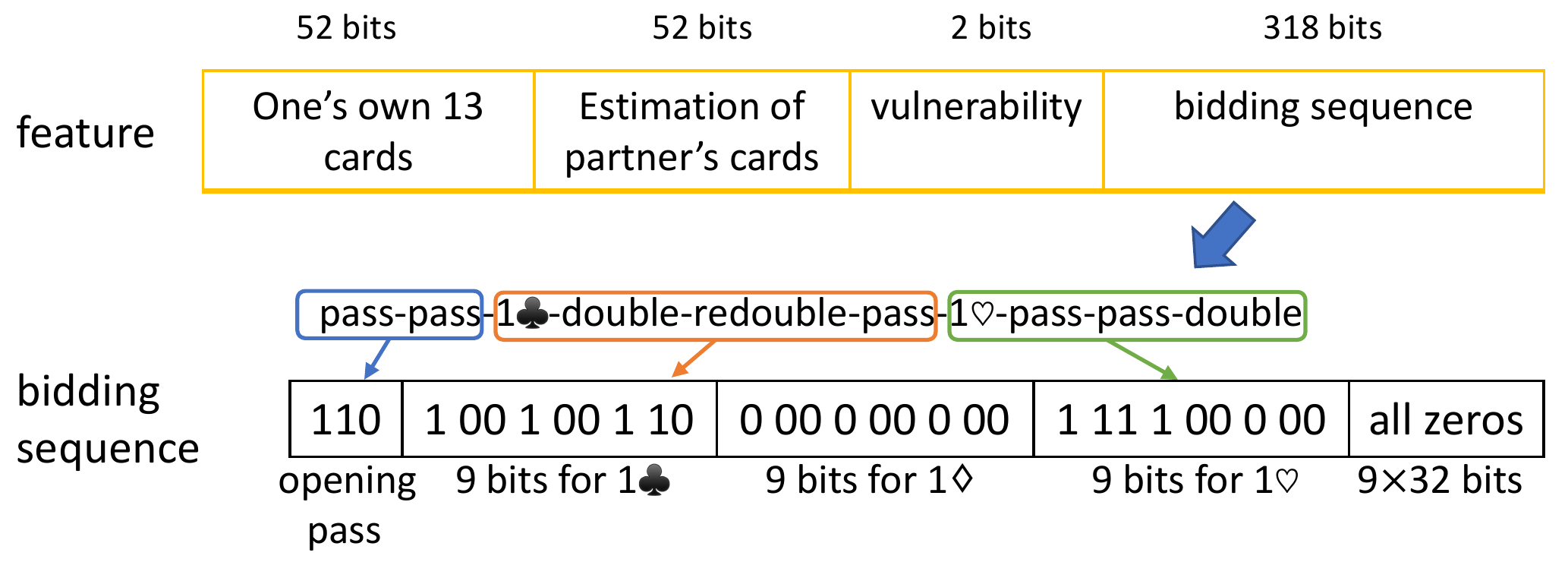}
	\caption{Compact representation of an example bidding sequence. \label{fig:compact}}
\end{figure*}

Figure \ref{fig:compact} shows an example of the representation of the compact feature with bidding sequence 
\begin{equation}
\mbox{pass-pass-1$\clubsuit$-double-redouble-pass-1$\heartsuit$-pass-pass},\nonumber
\end{equation}
where a ``1'' in position $i$ indicates that the $i$-th bid in the possible maximal bidding sequence is called. We do not need to represent the player identity of each historical bid because the bids are called by players one by one in a clockwise manner and the player to bid can correctly match observed bids to corresponding players directly from the bidding sequence.

\section{Learning Algorithms}
\label{sec:lerning_algorithms}
We introduce the learning algorithms in this section, which combines supervised learning from expert data and reinforcement learning from self-play.

The expert data are collected from the Vugraph Project \footnote{Vugraph Project. \url{http://www.bridgebase.com/vugraph_archives/vugraph_archives.php}}, which contains more than 1 million games from expert-level tournaments of the past 20 years and keeps adding more data constantly. The information of each game recorded in the dataset includes players' cards, the vulnerability, the dealer, the bidding sequence, the contract, the playing procedure, and the number of tricks won by the declarer.      

Since the PNN takes the output of the ENN as a part of its features, we first train the ENN based on the tuples of $(c_p, v, h, c_{p^+})$ generated from the expert data. The ENN is a multi-class multi-label classifier because the label $c_{p^+}$ contains 13 ones and 39 zeros. The output layer of the ENN consists of 52 sigmoid neurons. We calculate the cross entropy of each neuron and then sum them together as the final loss. When the training of the ENN is finished, we use it to generate the features of the PNN by modifying each instance $(c_p, v, h, c_{p^+})$ in the dataset into
\begin{equation}
(c_p, v, h, \phi_\omega(c_p, v, h), b_p),
\end{equation}
where $b_p$ is the bid called by player $p$. The PNN is a multi-class single-label classifier with a softmax output layer of 38 neurons. 
%

After the SL procedure, we further improve the ability of our bidding system by RL through self-play. We randomly generate the deal, including the dealer, the vulnerability and the cards of each player, and then the self-play starts. There are two bidding systems in the RL phase, one for the target team to be improved and the other for the opponents. Each deal is played twice, with the target system playing at the $N$-$S$ and $E$-$W$ positions, respectively. Two players of a partnership use the same bidding system (ENN and PNN) and a player's bid is sampled from the distribution output by his/her PNN. The self-play ends when a bid is followed by three ``passes''. 

Note that the final score depends on both the contract of bidding and the playing result. It is time consuming to play each deal out either by humans or by some playing program. To address this problem, we use DDA to approximate the playing outcome, which computes the number of tricks taken by each partnership for each possible contract of a deal under the assumption of perfect information and optimal playing strategy. Since players can exchange information during bidding, the result of DDA is usually close to the real one especially for professional players. The biggest advantage of DDA is its speed, usually within several seconds using double dummy solvers, e.g., the Bridge Calculator \footnote{Bridge Calculator. \url{http://bcalc.w8.pl/}}. Given the DDA results and the contract, we can compute the two partnerships' duplicate scores based on the rule of DBS. The duplicate score $r$ for the target system is then used to update the parameters $\theta$ of its PNN $\sigma_\theta(\cdot)$ according to the following equation:
\begin{equation}
\label{eq:REINFORCE_grad}
\theta \leftarrow \theta + \alpha r\frac{1}{M}\sum\limits_{i=1}^M\nabla_\theta\log(\sigma_\theta(b_i|s_i)),
\end{equation}
where $\alpha$ is the learning rate, $M$ is the number of bids called by the target PNN, $b_i$ and $s_i$ correspond to the $i$-th sampled bid and feature vector of $\sigma_\theta(\cdot)$, respectively, and $\sigma_\theta(b_i|s_i)$ is the probability of calling $b_i$ given the input $s_i$. The loss function of the ENN $\phi_\omega$ in the RL phase is the same with that in the SL procedure. 

We train the ENN and PNN simultaneously in the RL. The complete process is depicted in Algorithm \ref{alg:DRL}. To improve the stability of the training, we use a mini-batch of 100 games to update the parameters. Furthermore, following the practice of \cite{silver2016mastering}, we maintain a pool of opponents consisting of the target bidding systems in previous iterations of the RL. We add the latest bidding system into the pool every 100 updates and randomly select one for the opponents at each mini-batch of self-plays.

\begin{algorithm}
	\caption{Reinforcement learning for ENN and PNN\label{alg:DRL}}
	\KwIn{The ENN $\phi_\omega(\cdot)$ and PNN $\sigma_\theta(\cdot)$ from SL, and a set of randomly generated games ($G$)\;}
	\KwOut{The improved ENN and PNN from RL\;}
	Initialize the target bidding system with $\phi_\omega(\cdot)$ and $\sigma_\theta(\cdot)$\;
	Put $[\phi_\omega(\cdot), \sigma_\theta(\cdot)]$ in the empty opponent pool $O$\;
	\For {$\mbox{mini-batch}~~i = 1,2,\ldots, \frac{|G|}{100}$}
	{
		$\Omega_1\leftarrow \emptyset$\;
		$\Omega_2\leftarrow \emptyset$\;
		Randomly select a bidding system from $O$ for opponents\;
		\For {$\mbox{episode}~~j = 1,2,\ldots,100$}
		{
			Use the deal $c_{ij}\in C$ to initialize each player's cards, the vulnerability and the dealer\;
			\For {pos $=N$-$S$, $E$-$W$}
			{
				The target bidding system plays at position $pos$\;
				Let the four players bid until the end\;
				Use DDA to calculate the duplicate score $r$ for the target partnership\;
				Save the inputs, bids and $r$ of the target PNN in $\Omega_1$\;
				Save the inputs and corresponding cards of the target ENN in $\Omega_2$\;
			}
		}
		Update $\theta$ based on $\Omega_1$ and update $\omega$ based on $\Omega_2$\;
		Save $[\phi_\omega(\cdot), \sigma_\theta(\cdot)]$ in $O$ every 100 mini-batches\;
	}
\end{algorithm}

\section{Experimental Results}
\label{sec:experiments}
We conduct a set of experiments to evaluate our bidding system. We first give an overview of expert data and compare the DDA result with that of expert playing in the dataset to demonstrate that DDA is a good approximation of expert playing process and estimating the partner's cards is much more important than inferring opponents' cards. Next we present the detailed evaluation on the performance of the ENN and PNN. Finally, we test the strength of our bidding system.

\subsection{Expert Data Overview and DDA Evaluation}
\label{sec:DDA}

\begin{figure*}[h]
	\centering
	\subfigure[Contract distribution with suits.]{
		\label{fig:contract_dist}
		\includegraphics[width=75mm]{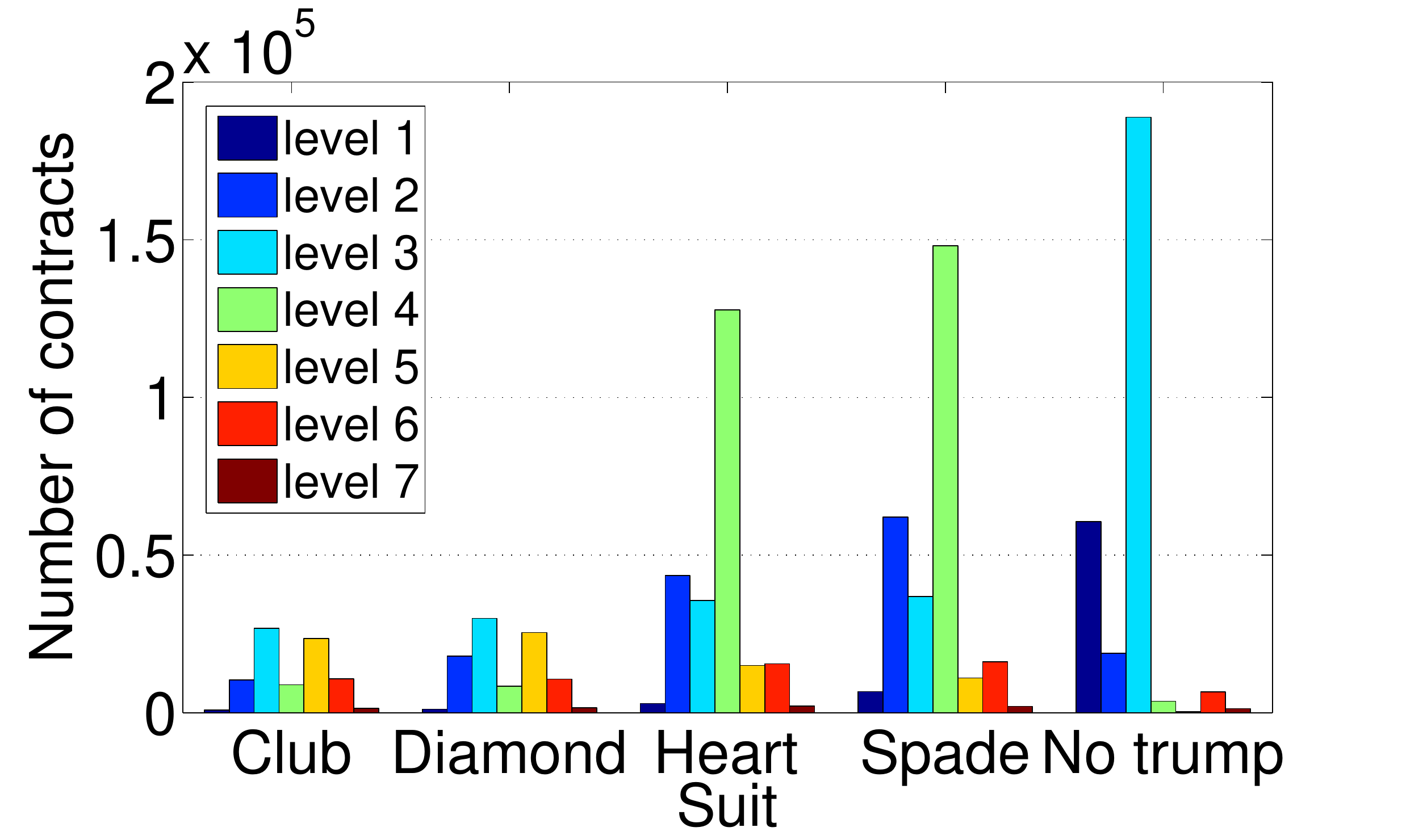}
	}
	\subfigure[Contract distribution with levels.]{
		\label{fig:contract_dist_level}
		\includegraphics[width=75mm]{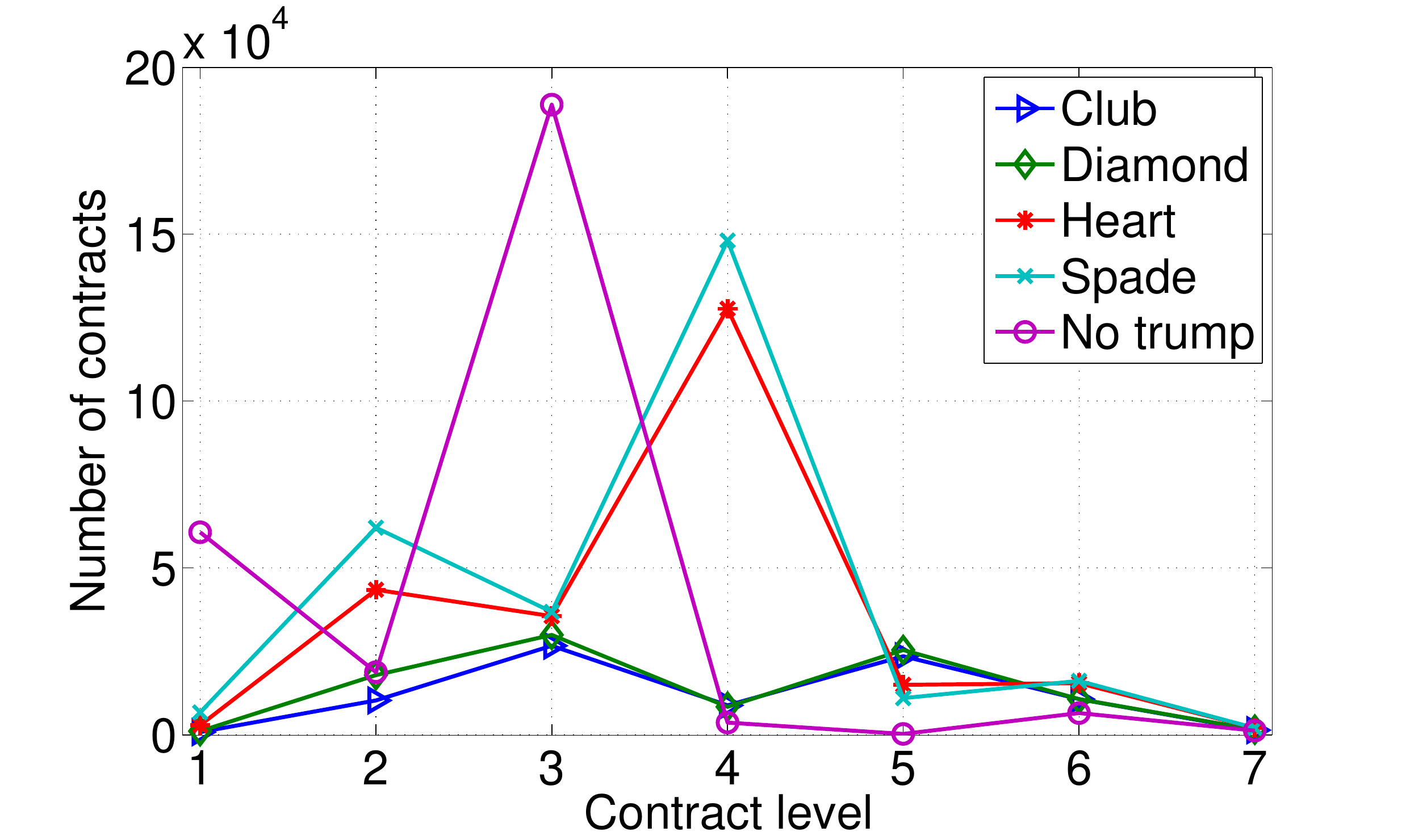}
	}
	\subfigure[Bidding length distribution.]{
		\label{fig:bidding_length}
		\includegraphics[width=75mm]{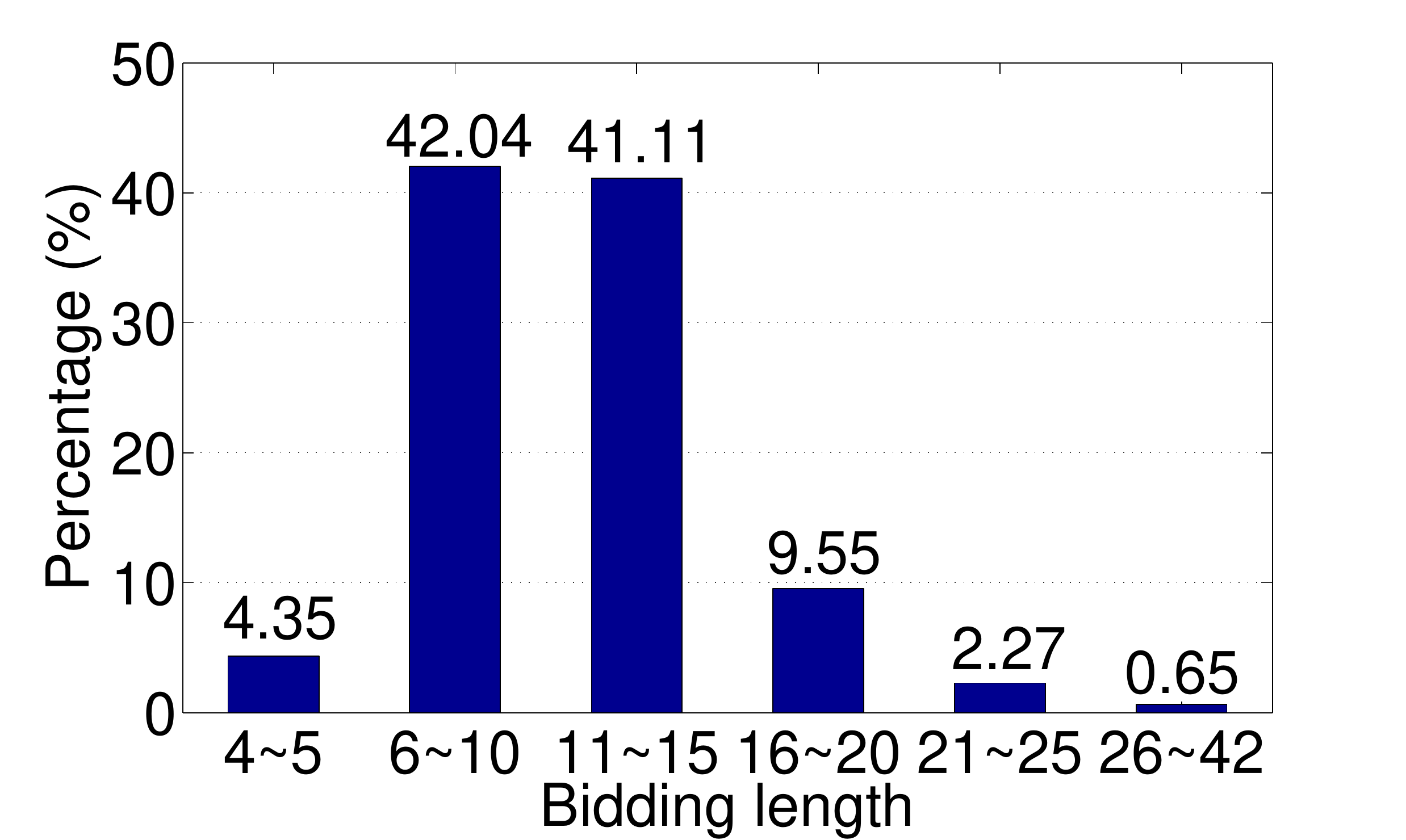}
	}
	\subfigure[Average number of overtakes.]{
		\label{fig:overtake_dist}
		\includegraphics[width=75mm]{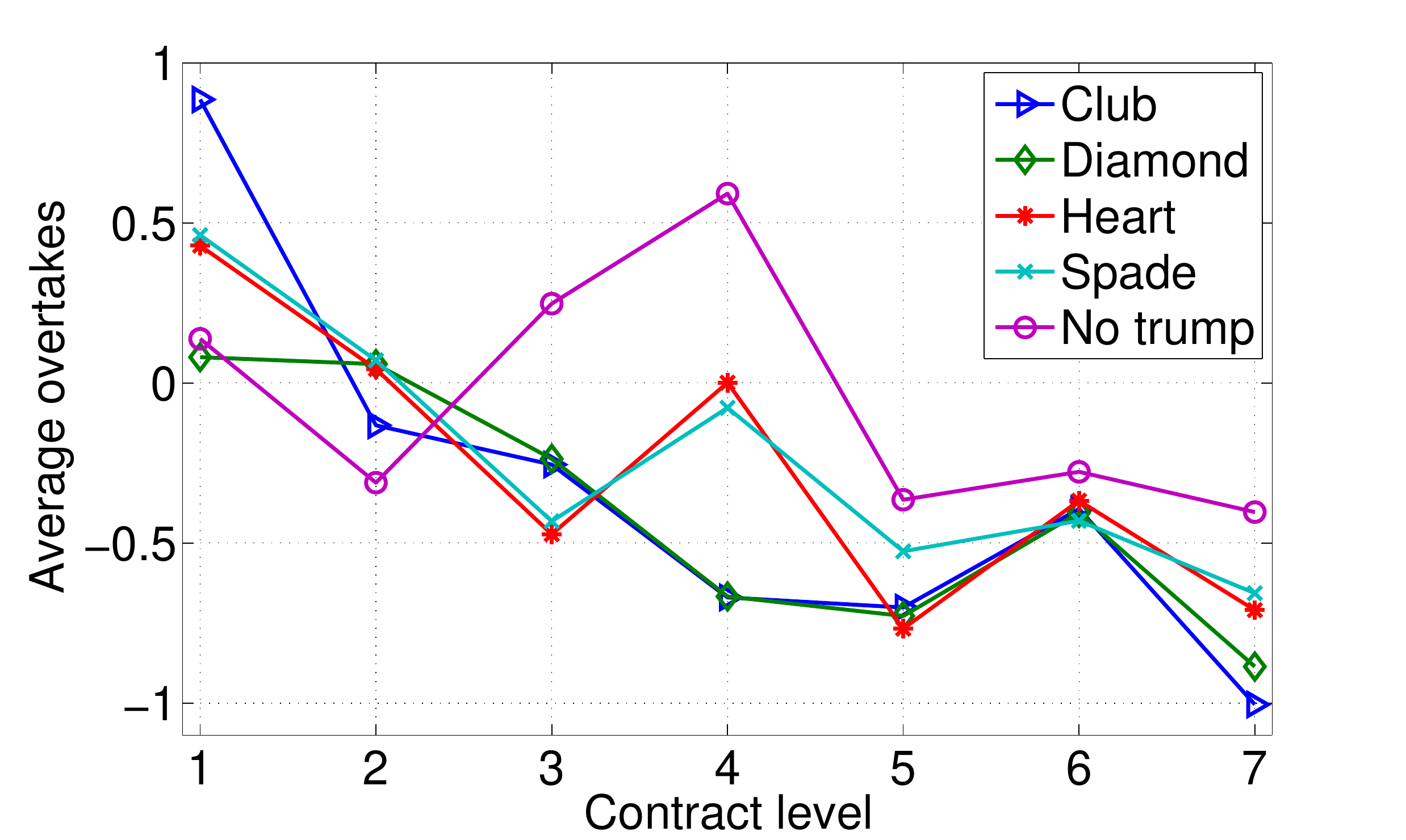}
	}
	\caption{Expert behaviors analysis. \label{fig:expert_behavior}}
\end{figure*}

The expert data include the information of the bidding and playing processes, the declarer, the contract and the number of tricks won by the declarer. There exist some noise in the data, e.g., incomplete records and mismatch between the tricks and the playing process. After filtering the noisy data, we finally get about 1 million games. 

The numbers of different contracts in the dataset are plotted in Figure \ref{fig:contract_dist}, from which we see that the major-suit and no-trump contracts are more preferred by experts. This observation is consistent with the DBS system, where the basic scores of making the major-suit and no-trump contracts are higher than those of minor-suit contracts. For example, basic scores of making $2NT$, $2\spadesuit$ and $2\clubsuit$ are 120, 110 and 90, respectively. Besides, we see from Figure \ref{fig:contract_dist_level} that most of the major-suit and no-trump contracts are at the level of 4 and 3, respectively, which is because contracts with level 4 of major suits and level 3 of no-trump suits constitute the ``game'' contracts and making them is worth 250 bonus scores. The distribution of lengths of bidding sequences in the data are depicted in Figure \ref{fig:bidding_length}, which indicates that most of the biddings run for 6$\sim$15 rounds. We see from Figure \ref{fig:overtake_dist} that the over takes with levels greater than 4 are negative, which means that it is difficult to make those contracts.

\begin{figure}[h]
	\includegraphics[width=75mm]{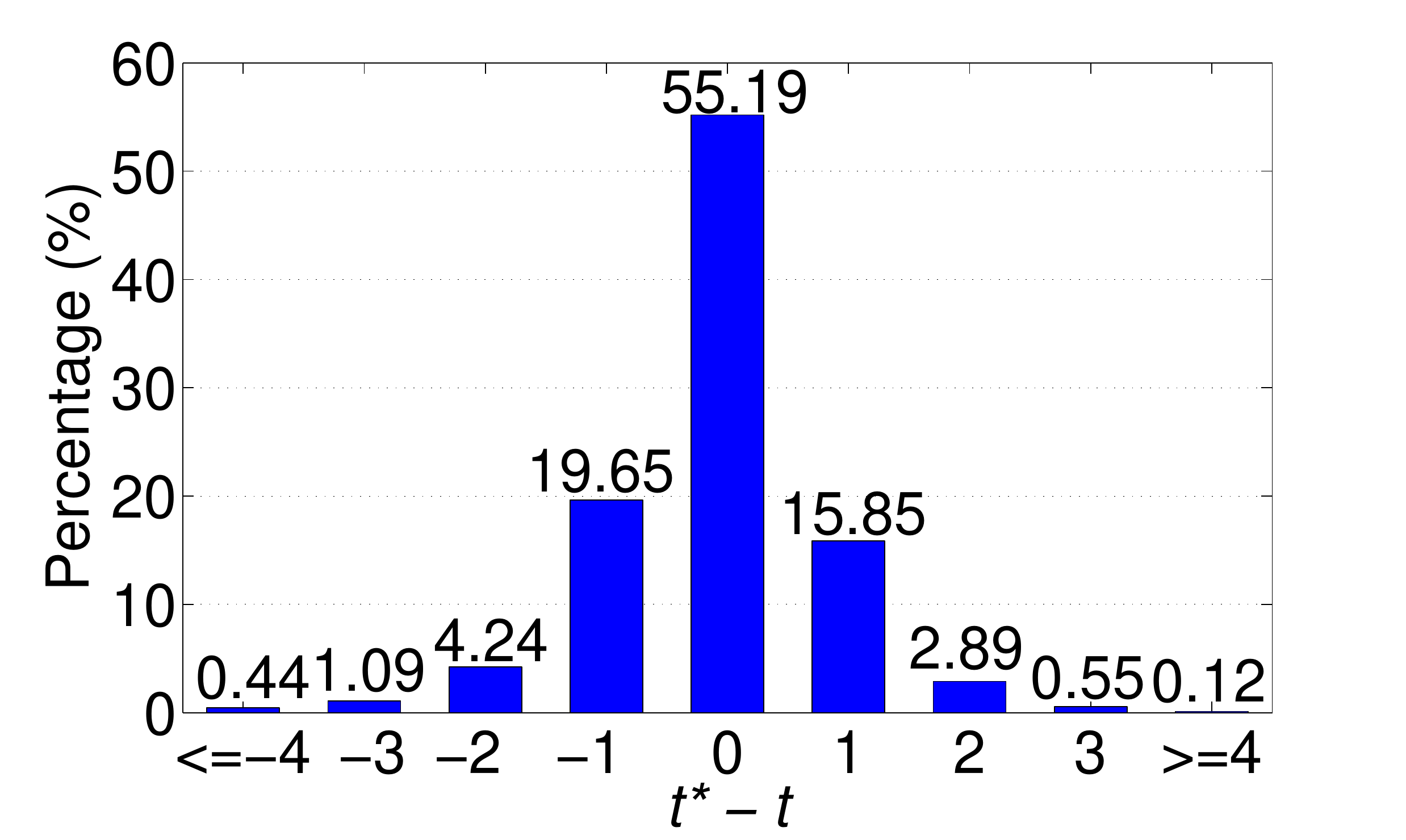}
	\caption{DDA evaluation. \label{fig:gap_dist_whole}}
\end{figure}

For each game, we use DDA to compute the declarer's tricks $t^*$ and calculate the gap, $t^* - t$, between the DDA's result and the real tricks $t$ in the data. The declarer's partnership can win at most 13 tricks in one deal and thus the range of the gap is $[-13, 13]$. Figure \ref{fig:gap_dist_whole} shows the distribution of the gap. As can be seen, more than 90$\%$ of the gaps 
lie in $[-1, 1]$ and $55.19\%$ of the gaps are equal to zero, which implies that DDA is a very good approximation to expert playing.

\begin{figure}[h]
	\includegraphics[width=75mm]{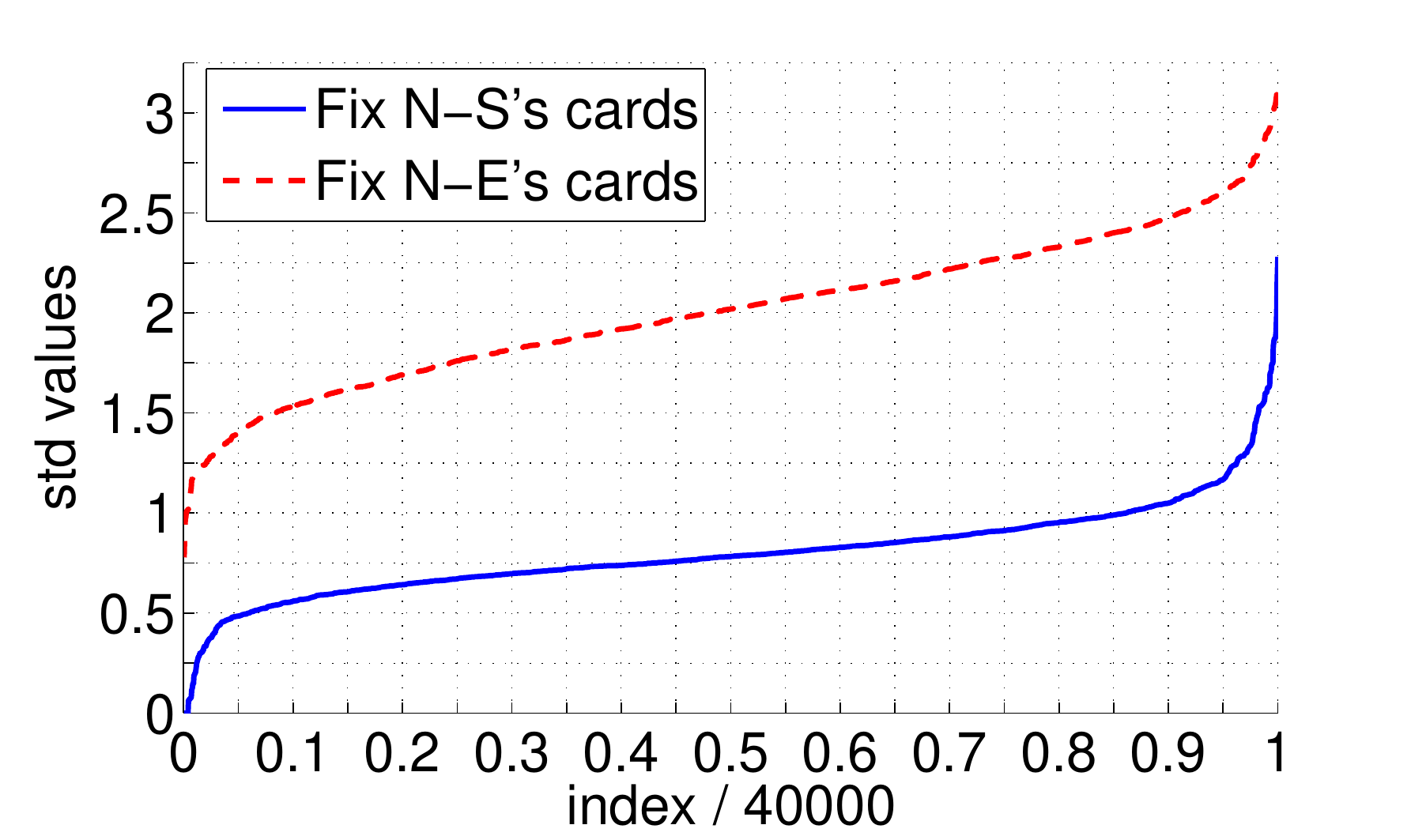}
	\caption{Importance comparison of estimating partner's and opponents' cards. \label{fig:demonstration}}
\end{figure}

Based on the DDA results, we demonstrate that 1) it is important to estimate our partner's cards for finding the optimal contract and 2) the card distribution in opponents' hands has little effect on the DDA results. Since there are 4 possible declarers and 5 possible trump suits, given the initial cards of four players, DDA outputs a $4\times5$ double dummy table (DDT), containing the numbers of tricks each possible declarer can win with each possible contract suit. The optimal contracts of the two partnerships can then be calculated based on the DDT and the DBS system. Thus, we just need to show that, given one's own cards, 1) the DDTs with different partner's cards are of high divergence and 2) the DDTs with different opponents' cards are similar. We use player $N$ as an example in the experiments. For demonstration 1, 2 thousand decks of initial cards for players $N$ and $E$ are randomly generated. For each deck, we use DDA to compute the DDTs of 1 thousand different initial cards of players $S$ (note that once the cards of $N$, $E$ and $S$ are given, the cards for $W$ can be obtained directly) and then calculate the standard deviation (std) of each declarer-suit pair over the 1 thousand DDTs. Finally we get $2000\times 4 \times 5 = 40000$ std values, which are indexed from the smallest to the largest and plotted in Figure \ref{fig:demonstration}. The second demonstration uses the similar method, except that 2 thousand decks of initial cards are generated for players $N$ and $S$, and then randomly sample 1 thousand $E$'s cards. The standard deviation caused by modifying $p$'s cards indicates how relevant the DDA result is to $p$’s cards. The lower the standard deviation is, the weaker the relevance between the result and $p$'s cards are, and the less important $p$’s cards are to the result. We see from Figure \ref{fig:demonstration} that about $90\%$ of the std values of different partner's cards are greater than 1.5 and $55\%$ of them are even greater than 2, while about $90\%$ of the std values of different opponents' cards are less than 1 and about half of them are less than 0.75, which are consistent with our expectation.

\subsection{Evaluation of ENN and PNN}
We first evaluate the fitting accuracy of the ENN in expert data. We generate more than 12 million training instances from the 1 million games, 70\% of which are used for training, 10\% for validation and 20\% for test. To increase the divergence of these dataset, the instances from a single game are put in the same dataset (training, validation or test). 

\begin{figure}[h]
	\includegraphics[width=75mm]{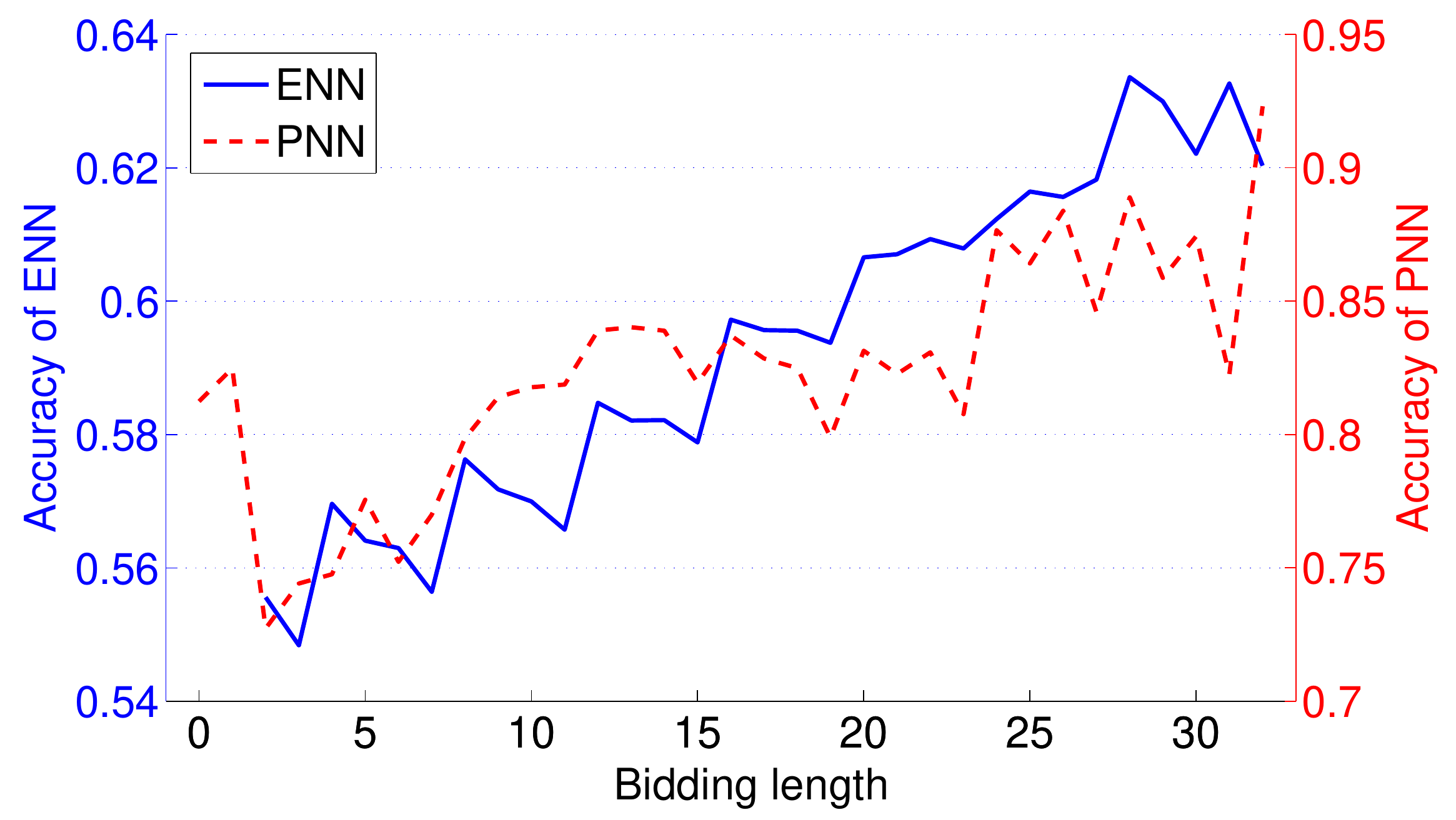}
	\caption{Accuracies with different bidding lengths. \label{fig:ENN_PNN_accuracy_length}}
\end{figure}

\begin{figure*}[h]
	\centering
	\subfigure[Accuracy and recall of ENN.]{
		\label{fig:ENN_accuracy_recall}
		\includegraphics[width=75mm]{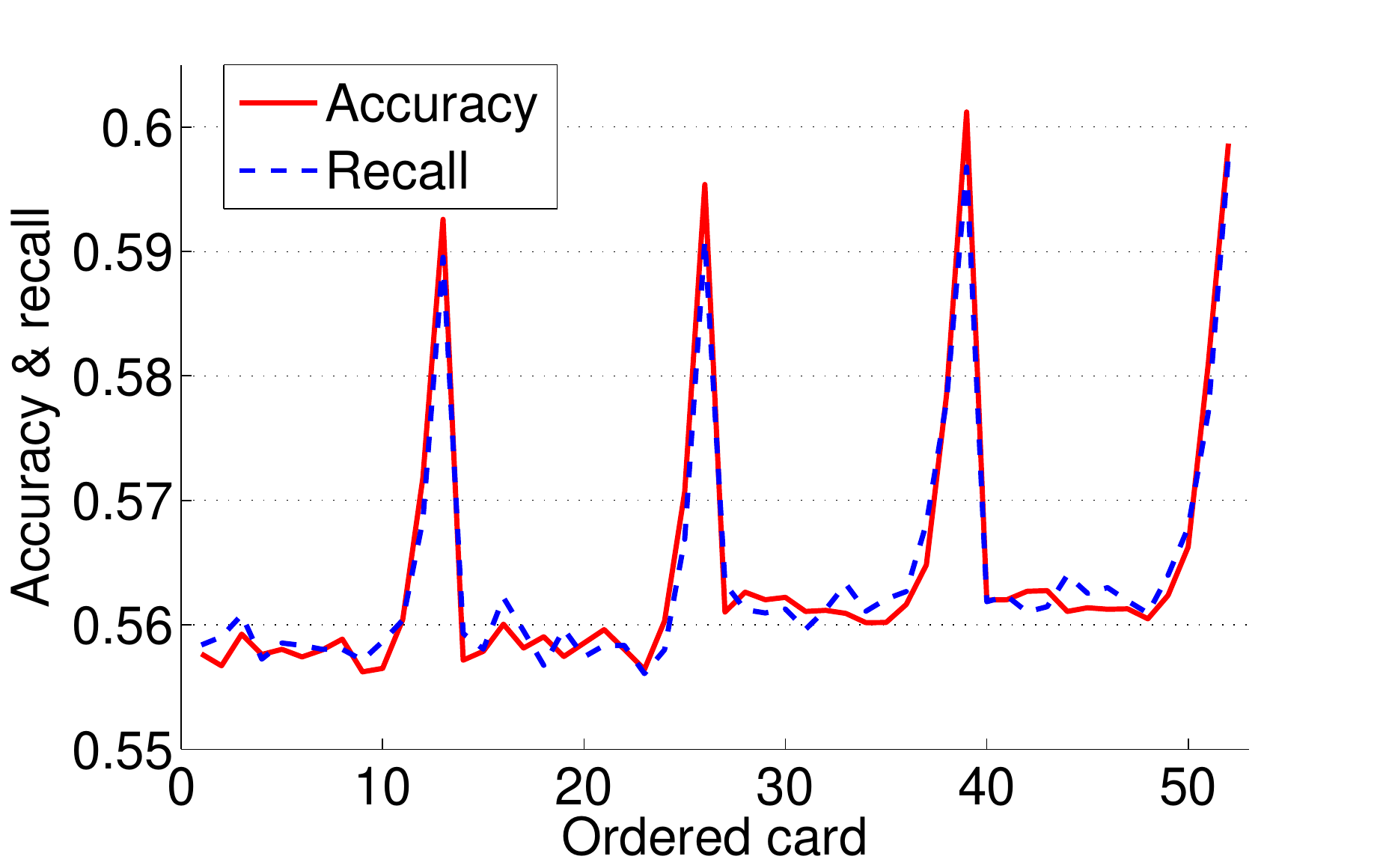}
	}
	\subfigure[Accuracy and recall of PNN.]{
		\label{fig:PNN_accuracy_recall}
		\includegraphics[width=75mm]{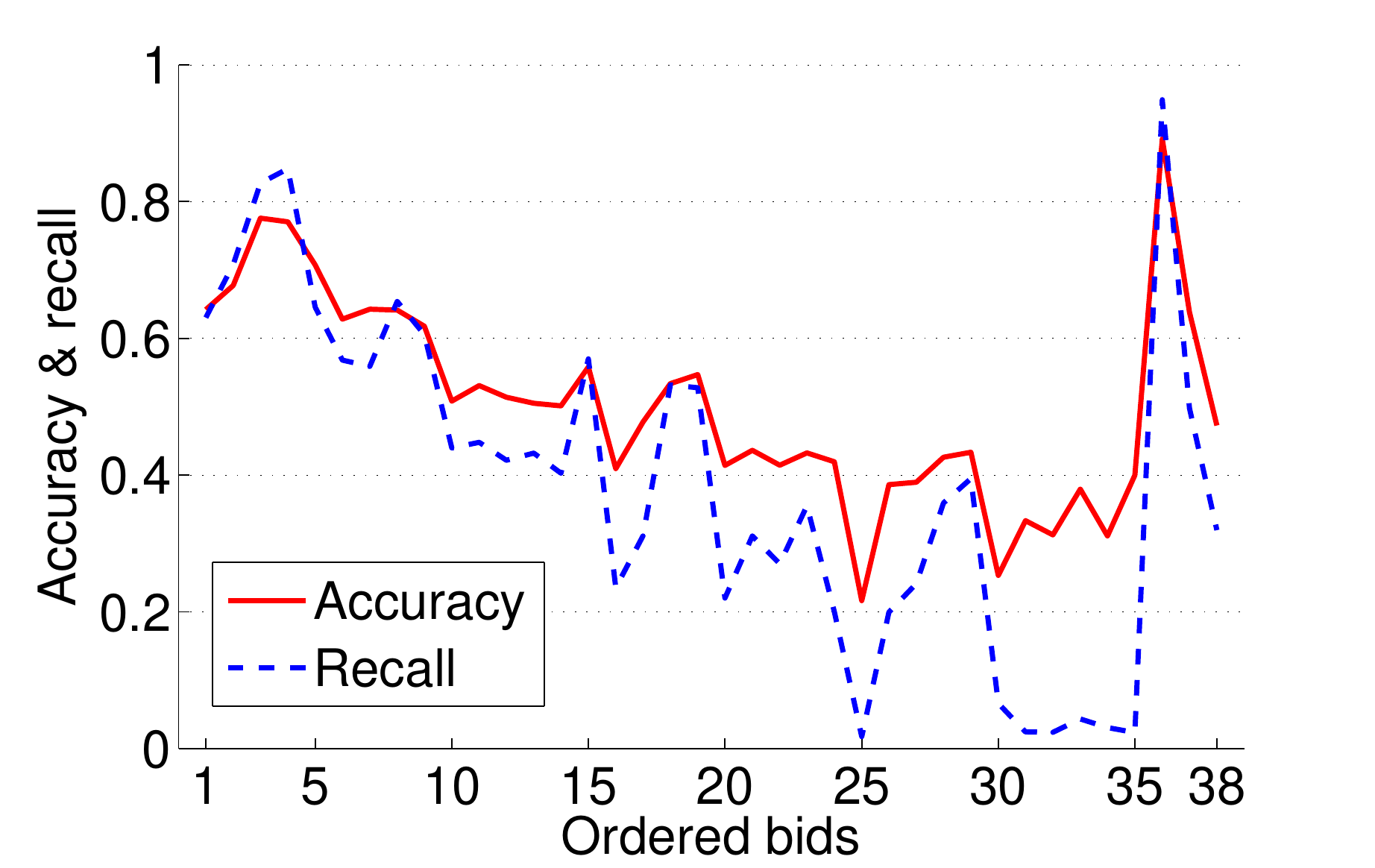}
	}
	\caption{Fitting accuracy of ENN and PNN. \label{fig:fitting_accuracy}}
\end{figure*}

The ENN uses a fully connected neural network with 8 layers, each hidden layer of which has 1500 neurons. Besides, we add an  extra skip connection every two layers for the network. To evaluate the accuracy of the ENN, cards with the highest 13 probabilities are selected as predictions of the ENN and then the accuracy is equal to the number of correct predictions divided by 13. Note that the PNN takes the 52 probabilities output by the ENN as input, but not the 13 highest-probability cards. The average accuracies of the ENN with different bidding length on the test dataset are shown in Figure \ref{fig:ENN_PNN_accuracy_length}. We see that the accuracies increase gradually. That is, the more actions we observe from our partner, the more accurately we can estimate his/her cards. The average accuracy and recall of each card is depicted in Figure \ref{fig:ENN_accuracy_recall}. It implies that the accuracy and recall of card of ``A'' are higher than other cards and the fitting performance on major suits is slightly better than that on minor suits. Because the number of possible partner's hand is $8\times10^9$, it is very difficult to get a high accuracy with very limited observations. In fact, in most situations, it is not necessary to know the exact cards of the partner, for example, to make a specific contract, one may just want to confirm that his/her partner holds at least 4 minor-suit cards, i.e., the sum of the probabilities of the ENN for minor suits is not less than 4, no matter which minor-suit card is held. Note that the outputted card distribution of the ENN is directly inputted to the PNN and thus such information can be used by the PNN for decision making.

The datasets for the PNN are generated based on the ENN. The PNN consists of a fully connected neural network of 10 layers with skip connection and each hidden layer has 1200 neurons. The accuracies of the PNN for predicting experts' moves with different bidding lengths are shown in Figure \ref{fig:ENN_PNN_accuracy_length}. We learn that the accuracies of the two partnerships' first bids are higher, which is because the rules for opening bids and corresponding responses are relatively well-defined. Overall, the accuracy also increases with the number of observed bids. The accuracies and recalls for different bids are shown in Figure \ref{fig:PNN_accuracy_recall}, where the first 35 indexes correspond to the 35 ordered contract bids and the last three (36, 37, 38) represent ``pass'', ``double'' and ``redouble'' respectively. The results indicate that the ``pass'' action is easy to predict because both the accuracy and recall are high. In fact, more than $50\%$ of the bids in the expert data are ``pass''. Besides, we see that the PNN performs better at low-level bids than at high-level bids, because the bidding usually begins with low-level bids and thus we have more data for training the PNN on them.

\begin{figure}[h]
	\centering
	\includegraphics[width=75mm]{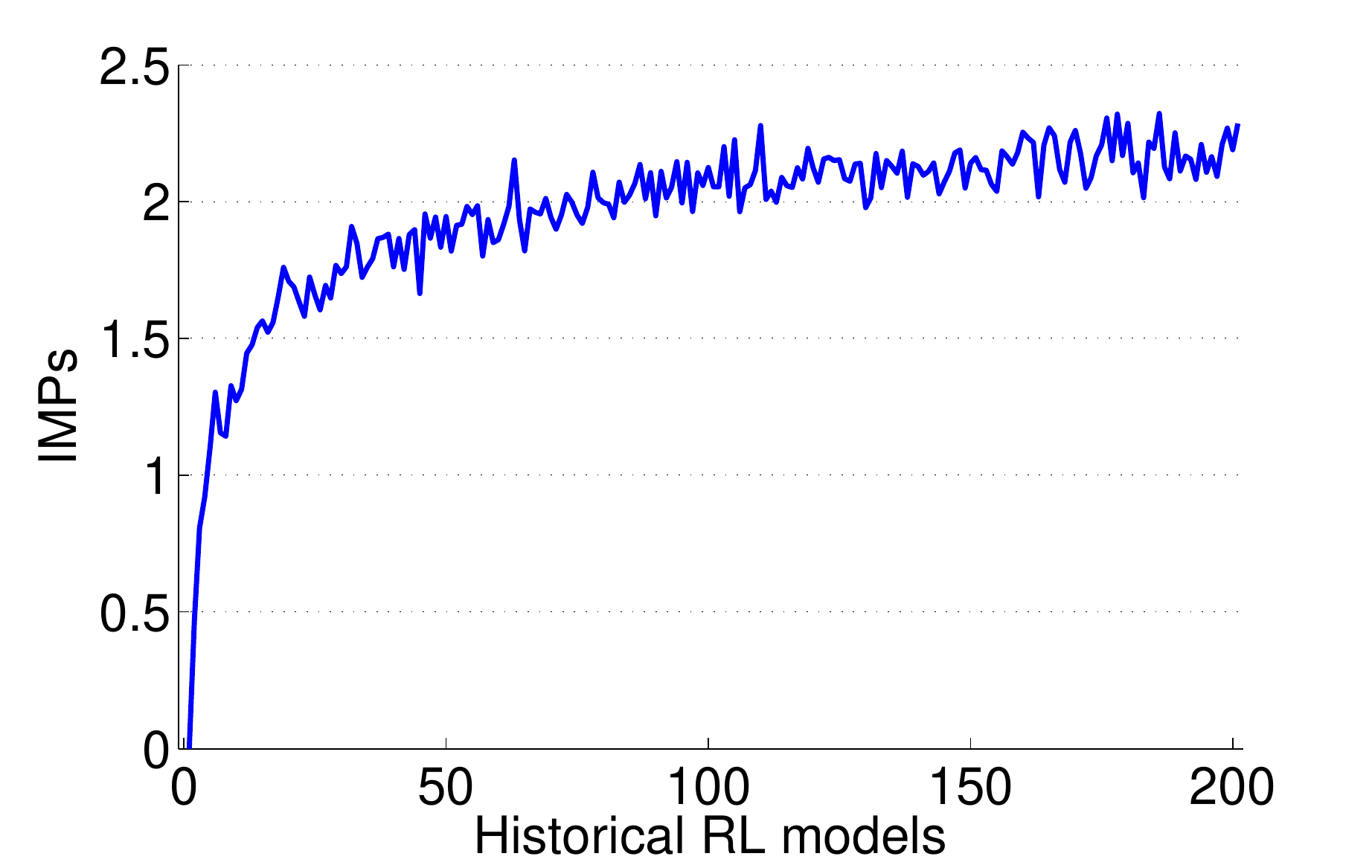}
	\caption{RL training curve. \label{fig:RL_curve}}
\end{figure}

Next, we evaluate the improvements of the bidding system in the RL procedure. We randomly generate 2 million deals and use Algorithm \ref{alg:DRL} to train the ENN and PNN. To distinguish with the different networks, we call the ENN (PNN) after the SL the SL-ENN (SL-PNN). Similarly, we use RL-ENN (RL-PNN) to denote the networks after the training of the RL. The opponent pool consists of the historical versions of the RL-ENN and RL-PNN in the RL. To evaluate the performance of the algorithm, we compare the historical networks with the initial SL networks through bidding competition over 10 thousand random deals. The average IMPs got by these RL networks in the competition is plotted in Figure \ref{fig:RL_curve}. As can be seen, the strength of the bidding system is significantly improved in the RL.
\begin{table*}[h]
	\centering
	\caption{Bidding strength comparison \label{tab:bidding_strength}}
	\begin{tabular}{cccccc}
		\hline
		& SL-PNN & RL-PNN & SL-PNN+ENN & RL-PNN+ENN & Wbridge5 \\
		\hline
		SL-PNN & N/A & -8.7793 & -5.653 & -9.2957 & -- \\
		RL-PNN & 8.7793 & N/A & 2.1006 & -1.0856 & --\\
		SL-PNN+ENN & 5.653 & -2.1006 & N/A & -2.2854 & --\\
		RL-PNN+ENN & 9.2957 & 1.0856 & 2.2854 & N/A & 0.25 \\
		\hline
	\end{tabular}
\end{table*} 

\subsection{Performance of the Bidding System}
To show the importance of the ENN, we trained another policy neural network without consideration of the partner's cards, which has the same structure with the PNN except that the feature of $\phi_\omega(\cdot)$ is removed. We use the notations of SL-PNN (RL-PNN) and SL-PNN+ENN (RL-PNN+ENN) to denote the bidding systems built with the SL (RL) version of a single PNN and the SL (RL) version of the PNN plus ENN in this subsection. The performances of different bidding systems are shown in Table \ref{tab:bidding_strength}, where the IMPs are in view of the row bidding systems and are averaged over 10 thousand random deals when both teams are network-based systems. 

We see that the RL bidding systems are stronger than SL systems and even the RL-PNN can beat the SL-PNN+ENN by 2.1006 IMPs, which implies that the RL can significantly improve the bidding system. The strongest RL-PNN+ENN beats RL-PNN by 1.0856 IMPs, which indicates that the ENN is a key component of our bidding system. The comparison with Wbridge5 is manually tested on 64 random boards because there is neither code nor command line interface for Wbridge5. We just compare the bidding ability with Wbridge5 and the scores are also computed with DDA. Wbridge5 implements many bidding rules, e.g., Weak Two, Strong 2D and Unusual 2NT, which can be selected by players in the ``Options'' of Wbridge5. The comparison results indicate that our best RL-PNN+ENN system is stronger in bidding, with a positive average IMP of 0.25 over Wbridge5. It is claimed that a 0.1 IMP improvement is significant for bidding strength \cite{ventos2017boosting}.

\section{Conclusion and Discussion}
In this paper, we designed a neural network-based bidding system, consisting of an estimation neural network (ENN) for inferring the partner's cards and a policy neutral network (PNN) to select bids based on the public information and the output of the ENN. Experimental results indicate that our system outperforms the top rule-based program -- Wbridge5. 

Contract bridge is a good testbed for artificial intelligence because it is one of the most difficult card games, involving large state space, competition, cooperation and imperfect information. Our methods can be applied to other games. Specifically, the feature representation method in our paper provides a general idea to efficiently encode the action history in a game where the maximal history length is finite. For example, the method can be applied to limit Texas Hold’em poker. Since the possible action sequences in each round and possible numbers of round in the game are finite, we can use a vector whose length is equal to the maximal action sequence to encode the action history, where a “1” in position $i$ means that the corresponding action is taken, while a “0” means not taken.

Besides, how to deal with private information of other players in imperfect-information games is a key problem. Although lack of theoretical support, our work experimentally shows that using a particular estimation component is effective. Since bridge is a very complex game with multiple players, imperfect information, collaboration and competition, the experimental evidence can motivate the method to be applied to other games, e.g., multi-player no-limit Texas Hold’em poker and majhong.

For future work, first we will further improve the strength of our system. Second, we will develop a computer program for bridge and open to the community for public test. Third, we will ``translate'' the network-based system to a convention to play with humans.

\begin{acks}
This paper is supported by 2015 Microsoft Research Asia Collaborative Research Program.
\end{acks}

\bibliographystyle{ACM-Reference-Format}  
\bibliography{bridge}  

\end{document}